\documentclass[letterpaper, 10 pt, conference]{ieeeconf}  
\IEEEoverridecommandlockouts
\usepackage{amsmath,amssymb,amsfonts}
\usepackage{algorithm}
\usepackage{algpseudocode}
\usepackage{graphicx}
\usepackage{makecell}
\usepackage{textcomp}
\usepackage{xcolor}
\usepackage{pifont}

\let\labelindent\relax
\usepackage{enumitem}

\usepackage{multirow} 
\usepackage{tabularx}
\usepackage{booktabs}
\usepackage{siunitx}

\usepackage{setspace}
\usepackage{epstopdf}

\usepackage{wrapfig}
\usepackage{mathtools}
\usepackage{interval}

\usepackage{soul}

\title{\LARGE \bf
Decoupling Ego-Motion from Target Dynamics via Dual-Interval Motion Cues for UAV Detection}

\author{Liuyang Wang and Feitian Zhang*
\thanks{All the authors are with the Robotics and Control Laboratory, Department of Robotics, School of Advanced Manufacturing and Robotics, and the State Key Laboratory of Turbulence and Complex Systems, Peking University, Beijing, 100871, China (liuyang.wang@stu.pku.edu.cn; feitian@pku.edu.cn). L. Wang is also with Great Bay University, Guangdong 523808, China.}
}

\begin{document}
\bstctlcite{IEEEexample:BSTcontrol} 

\maketitle
\thispagestyle{empty}
\pagestyle{empty}

\begin{abstract}
Object detection from Unmanned Aerial Vehicles (UAVs) is challenged by severe ego-motion, camera jitter, and large scale variations. While modern detectors perform well on static images, their direct application to UAV video often fails, particularly for small objects in dynamic scenes. Existing motion-based methods often rely on computationally expensive optical flow.
We propose a vision-only motion-guided detection framework that decouples target motion from camera-induced disturbances. Homography-based Global Motion Compensation (GMC) first aligns adjacent frames. We then introduce a Dual-Interval Motion Extraction strategy that captures both short-term and long-term motion cues. To integrate these cues, a lightweight Dual-Interval Motion Injection (DIMI) module enhances feature representations within a Feature Pyramid Network.
Experiments on the VisDrone-VID dataset demonstrate consistent improvements over the YOLOv8n baseline under severe ego-motion. Ablation studies further confirm the effectiveness of the dual-interval design and DIMI.
\end{abstract}


\section{Introduction}

\begin{figure}[!t]
    \centering
    \includegraphics[width=0.48\textwidth]{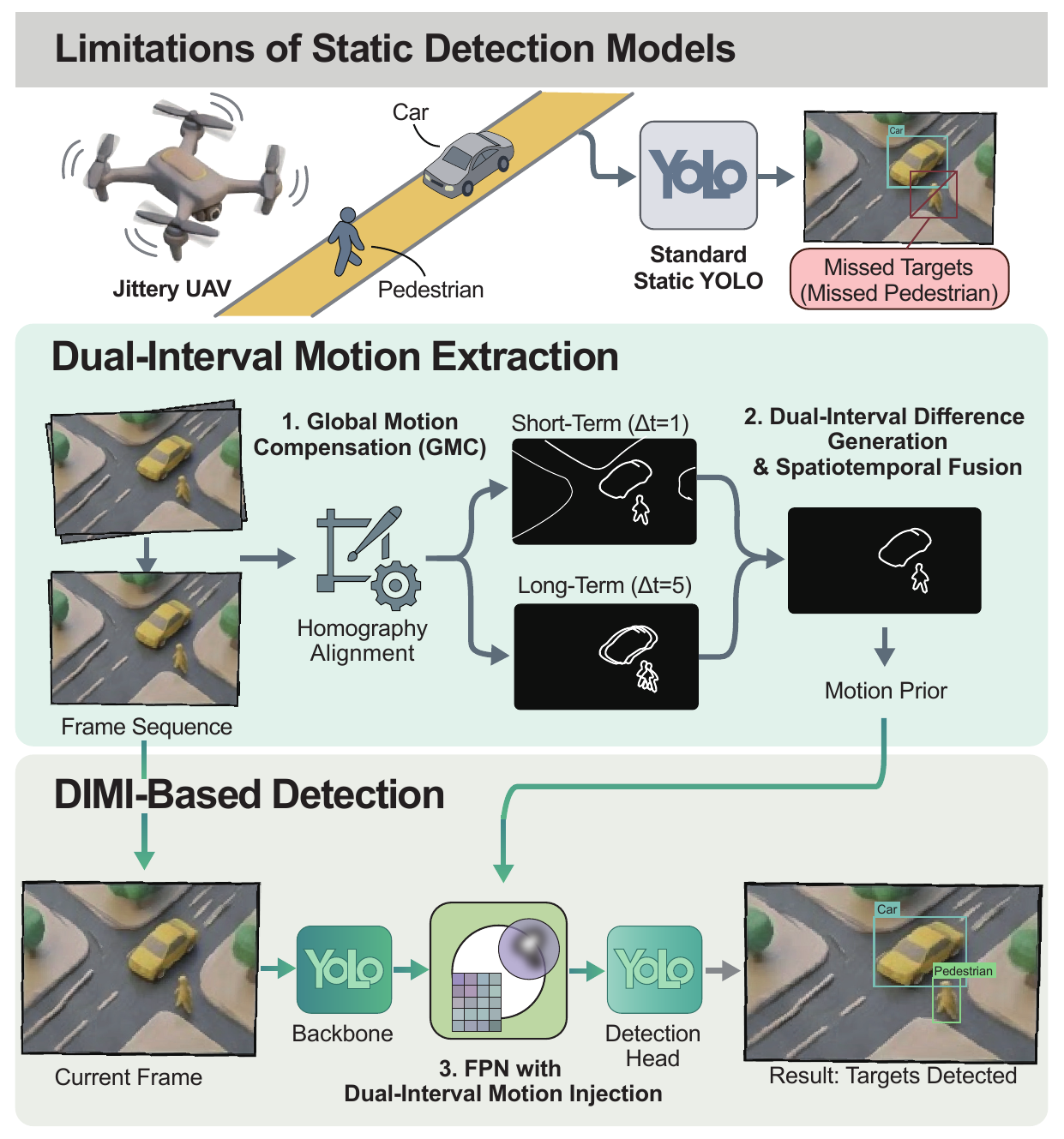}
    \caption{Illustration of the challenges in UAV object detection and the proposed solution. Top: Static frame-wise detectors often miss small or dynamic targets in UAV videos due to ego-motion and camera jitter. Middle: Global Motion Compensation (GMC) aligns frame sequences, while dual-interval temporal differencing extracts short- and long-term motion cues. Bottom: The Dual-Interval Motion Injection (DIMI) module utilizes the extracted motion cues to enhance target-aware feature representations and recover some targets missed by the baseline.}
    \label{fig:teaser}
\end{figure}

Unmanned Aerial Vehicles (UAVs) have become important platforms for infrastructure inspection, autonomous logistics, and environmental surveillance \cite{ollero2022past, ASMA, SIGN}. Robust and real-time object detection is central for UAV operation. Modern single-stage detectors, particularly the YOLO series, achieve a favorable balance between inference speed and detection accuracy on static images \cite{redmon2016yolo, yolov8_ultralytics}. 
However, directly applying these detectors to UAV video streams remains challenging due to severe ego-motion, camera jitter, and rapidly changing viewpoints \cite{bozcan2020auair, Kyrkou2020EmergencyNet}. Since frame-wise detectors process each image independently, valuable temporal information is discarded, often leading to degraded performance for objects in cluttered aerial scenes \cite{VisDrone, Perception}.

Existing motion-guided approaches typically rely on optical flow or frame differencing to extract motion cues \cite{fgfa}. Optical flow provides dense motion fields but introduces substantial computational overhead and sensitivity to illumination changes, limiting deployment on resource-constrained UAV platforms \cite{dosovitskiy2015flownet, Real-Time}. Conversely, frame differencing is lightweight but highly susceptible to motion and camera jitter \cite{yilmaz2006object}. Without proper alignment, ego-motion induces strong spurious responses that corrupt feature representations. Moreover, motion extraction based on a single time interval faces an inherent dilemma: short intervals capture exact current positions but are highly susceptible to ego-motion noise, whereas long intervals accumulate robust motion saliency but introduce severe spatial ghosting.

To address these challenges, we propose a vision-only, motion-guided object detection framework built upon YOLO, as illustrated in Fig.~\ref{fig:teaser}. 
Homography-based Global Motion Compensation (GMC) first aligns adjacent frames. We then introduce a Dual-Interval Motion Extraction strategy that leverages a mutual gating mechanism. By computing the consensus between short-term ($\Delta t=1$) localization and long-term ($\Delta t=5$) saliency, it successfully captures precise motion cues while filtering out both background noise and ghosting artifacts.
We further use a lightweight Dual-Interval Motion Injection (DIMI) module that embeds the extracted motion masks into the Feature Pyramid Network (FPN) as soft spatial attention, thereby enhancing target-related features across multiple scales.

The main contributions of this paper are threefold: 

\begin{itemize}[label={$\bullet$}, leftmargin=2em]
    \item We propose a unified vision-only motion-guided detection framework for UAV videos that integrates global motion compensation with dual-interval temporal differencing to robustly extract motion cues under severe ego-motion and camera jitter. 
    
    \item We design a lightweight DIMI module that incorporates motion masks into the FPN as soft spatial attention, improving the representation of objects with minimal computational overhead.

    \item We propose an asymmetric training-inference strategy that employs high-precision SIFT matching during offline training and lightweight ORB-based homography cascading during online inference, enabling real-time deployment on resource-constrained UAV platforms.
\end{itemize}

On VisDrone-VID, the proposed framework achieves 25.8 $\mathrm{AP}_{50}$ with a 3.01 M-parameter detector and runs at 36.8 FPS end-to-end on the Jetson Orin Nano.

\section{Related Work}

\begin{figure*}[t]
    \centering
    \includegraphics[width=0.9\linewidth]{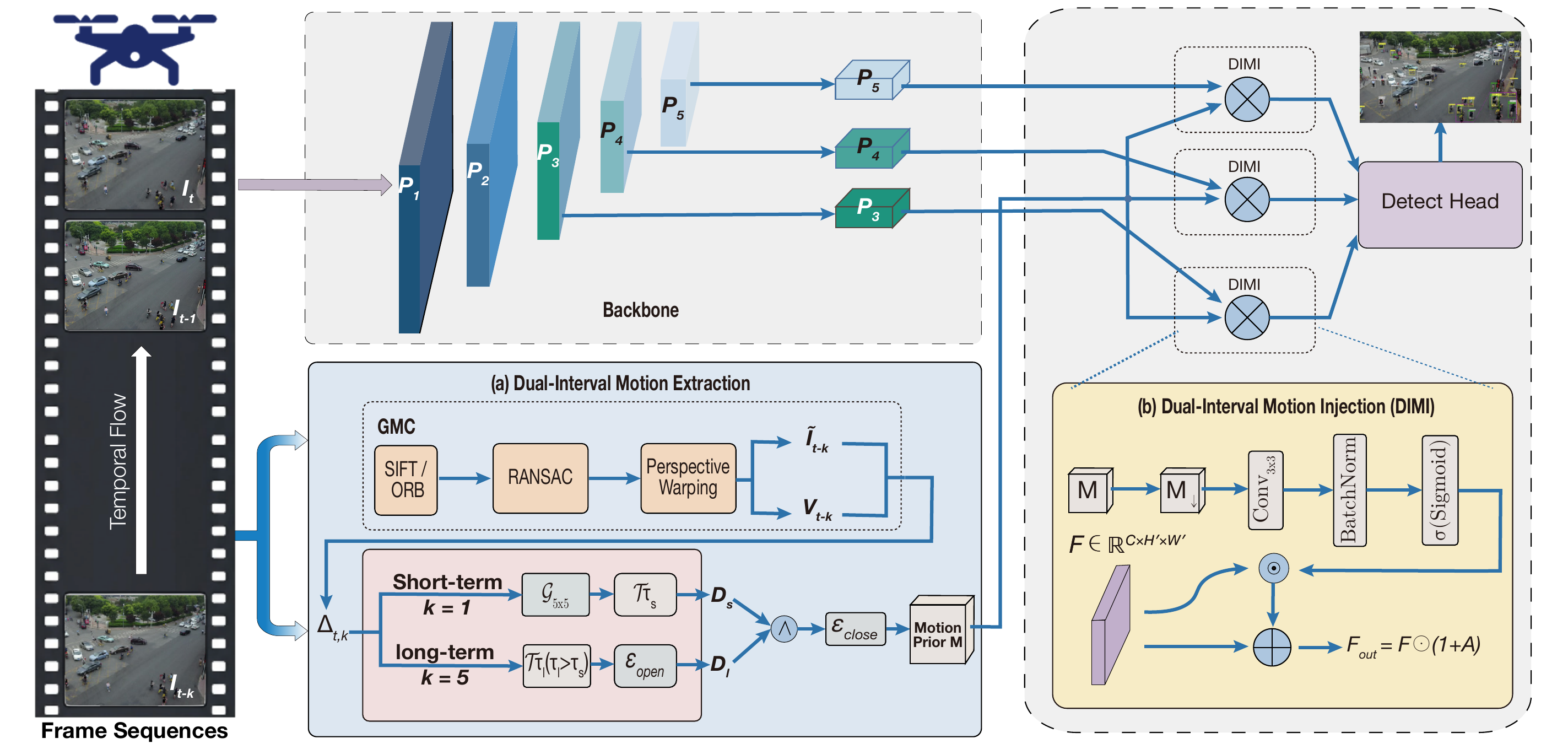}
    \caption{Overall architecture of the proposed motion-guided object detection framework. A YOLO backbone extracts multi-scale spatial features ($P_3, P_4, P_5$) from the current frame $I_t$. In parallel, the Dual-Interval Motion Extraction  module aligns historical frames ($I_{t-1}, I_{t-5}$) using Global Motion Compensation (GMC) and fuses short- and long-term temporal differences to generate a sparse motion mask $M$. The Dual-Interval Motion Injection (DIMI) module then injects the motion mask $M$ into the Feature Pyramid Network (FPN) via a lightweight attention mechanism to enhance target-aware feature representations before the final detection head.}
    \label{fig:Overview}
\end{figure*}

\subsection{Object Detection in UAV and Dynamic Environments}

Object detection methods are generally categorized into two-stage and single-stage paradigms. While two-stage detectors emphasize localization accuracy, single-stage architectures such as SSD \cite{SSD}, RetinaNet \cite{RetinaNet}, and the YOLO series \cite{yolov8_ultralytics} provide an effective balance between accuracy and real-time efficiency. More recently, transformer-based detectors such as DETR \cite{DETR} introduced global feature reasoning for end-to-end detection. These detectors have been widely evaluated on UAV benchmarks \cite{VisDrone, Det-Fly, AOT-C} and applied to aerial perception tasks \cite{AD-YOLO, ActiveClassification, 9519550, UEVAVD}.

 
\subsection{Motion-Guided Detection Under Dynamic Backgrounds}

To enhance robustness in dynamic environments, recent studies incorporate motion cues into object detection pipelines. 
Optical-flow-based methods explicitly model dense motion fields but are computationally expensive for edge deployment \cite{fgfa,dosovitskiy2015flownet}.

Motion cues have long been exploited in aerial and satellite video. In UAV videos, homography-based image registration has been combined with spatiotemporal differencing for moving-target detection \mbox{\cite{Minaeian}}. Three-frame differencing is used to locate moving targets in satellite video {\cite{Du2018}}, and accumulated differencing selects a longer temporal window for slow motion and a shorter window for fast motion {\cite{jimaging6040025}}. Two representative designs have recently been proposed for satellite video: DSFNet \mbox{\cite{DSFNet}} pairs a 2-D appearance backbone with a learnable 3-D convolutional stream that extracts motion implicitly from consecutive RGB frames, while the VISO benchmark \mbox{\cite{VISO}} introduces Accumulative Multi-Frame Differencing (AMFD), which averages the three difference masks formed from three neighbouring frames under an adaptive threshold to detect slow-moving objects.

Our front-end keeps two unequal causal intervals and intersects them, whereas AMFD averages three neighbouring-frame differences and uses the future frame $I_{t+1}$. The mask is an external prior for a single detector, whereas DSFNet adds a second, learnable 3-D stream.

Motion-guided detectors based on temporal differencing and motion fusion have demonstrated improved efficiency. For instance, MG-VTOD {\cite{yang2023video}} fuses motion-strength and appearance features after shallow convolution. GL-YOMO {\cite{GL-YOMO}} employs a cascaded global-local strategy, triggering a template-matching motion detector primarily when the static YOLO confidence is low. MA-YOLO {\cite{MA-YOLO}} performs detection on sparse keyframes and propagates the resulting boxes with precomputed H.264 motion vectors.

\subsection{Attention Mechanisms in Object Detection}
Attention mechanisms are widely adopted to improve feature representation and suppress background interference in object detection. 
Prior works incorporate spatial and channel attention to improve robustness against occlusion, illumination variation, and small-object ambiguity in UAV \cite{LAM-YOLO,YOLO-CAM}.

Despite these advances, most existing attention mechanisms (such as CBAM \mbox{\cite{CBAM}}) derive spatial attention endogenously by pooling internal feature channels. While specific works like Li et al. \mbox{\cite{Li2019}} have explored motion to guide attention for video salient object detection, our proposed DIMI module differs fundamentally in its input source.
Unlike methods that rely on endogenous feature attention, our DIMI module leverages exogenous, physically-grounded dual-interval motion priors to explicitly guide spatial attention. By integrating these cues into the FPN, the proposed module enhances motion-consistent regions across multiple scales through soft spatial attention.

\section{Method}
\subsection{Problem Formulation}

Given a UAV video stream $\mathcal{V} = \{I_t\}_{t=1}^T$, where $I_t \in \mathbb{R}^{H \times W \times 3}$ denotes the RGB frame at time step $t$, the objective of UAV-based object detection is to predict a set of bounding boxes and corresponding class labels $\mathcal{B}_t = \{(b_i, c_i)\}_{i=1}^N$ for each frame. 

Conventional static detectors formulate this task as an isolated spatial mapping $\mathcal{F}_\text{static}: I_t \rightarrow \mathcal{B}_t$, inherently discarding the motion cues present in $\mathcal{V}$. Conceptually, the observed image motion can be viewed as a superposition of target-induced and camera-induced components:
\begin{align}
    V_\text{obs} = V_\text{tgt} + \Delta_\text{ego},
\end{align}
where $V_\text{tgt}$ represents the target motion and $\Delta_\text{ego}$ denotes the ego-motion caused by UAV movement and camera jitter. Compensation of $\Delta_\text{ego}$ by a single homography assumes a planar scene or purely rotational camera motion.

To suppress camera-induced background motion and highlight target dynamics, we formulate a motion-guided detection paradigm $\mathcal{F}_\text{motion}: (I_t, \mathcal{M}_t) \rightarrow \mathcal{B}_t$, where $\mathcal{M}_t \in \{0, 1\}^{H \times W}$ denotes a sparse motion mask  that highlights target dynamics ($V_\text{tgt}$) while suppressing ego-motion disturbances ($\Delta_\text{ego}$).

\subsection{Framework Overview}

As illustrated in Fig.~\mbox{\ref{fig:Overview}}, the framework combines a single learnable detection network with a parameter-free motion front-end. The backbone receives only the 3-channel RGB frame and extracts multi-scale spatial features.
Dual-Interval Motion Extraction processes consecutive frame sequences, operating in parallel with the backbone. It intrinsically utilizes Global Motion Compensation (GMC) to align the background, followed by a dual-interval differencing strategy to generate a robust, sparse motion mask {$M$}.
DIMI uses this external mask to provide soft spatial attention in the FPN, enhancing the features before they enter the detection head.

\subsection{Dual-Interval Motion Extraction}\label{subsec3}


\textbf{Global Motion Compensation.} Let $I_t \in \mathbb{R}^{H \times W}$ denote the grayscale representation at time $t$, and $I_{t-k}$ denote a historical frame with temporal interval $k$. To align the background, we extract robust scale-invariant keypoints (e.g., SIFT \cite{2004Distinctive}) and estimate the geometric transformation matrices $H_{t, t-k}$ independently via RANSAC \cite{1981Random} to reject dynamic outliers. The reference frames are then warped into the coordinate system of $I_t$, formulated as
\begin{align}
    \tilde{I}_{t-k} = \mathcal{W}(I_{t-k}, H_{t, t-k}),
\end{align}
where $\mathcal{W}$ denotes the perspective warping operation. To prevent false motion responses at the image boundaries introduced by the geometric transformation, a valid region mask $R_{t-k} \in \{0, 1\}^{H \times W}$ is constructed by applying the identical warping $\mathcal{W}$ to an all-one matrix. The compensated difference map is subsequently computed as
\begin{align}
    \Delta_{t,k} = |I_t - \tilde{I}_{t-k}| \odot R_{t-k},
\end{align}
where $\odot$ denotes the Hadamard element-wise product.

\begin{algorithm}[t]
  \algdef{SE}[SUBALG]{Indent}{EndIndent}{}{\algorithmicend\ }%
  \algtext*{Indent}
  \algtext*{EndIndent}

  \algnewcommand\algorithmicinput{\textbf{Input:~}}
  \algnewcommand\algorithmicoutput{\textbf{Output:~}}
  \algnewcommand\algorithmicparameters{\textbf{Parameters:~}}
  \algnewcommand\Input{\State\algorithmicinput}%
  \algnewcommand\Output{\State\algorithmicoutput}%
  \algnewcommand\Parameters{\State\algorithmicparameters}%
  \algnewcommand{\LineComment}[1]{\State \(\triangleright\) #1}
  \caption{Dual-Interval Motion Extraction}\label{alg:dual_interval}
  \begin{algorithmic}[1]
    
    \Input
    \Indent
    \State $I_t$: Current grayscale frame
    \State $I_{t-1}, I_{t-5}$: Short-term and long-term reference frames
    \EndIndent

    \Output
    \Indent
    \State $M$: Robust spatiotemporal motion mask
    \EndIndent

    \Parameters
    \Indent
    \State $\tau_s, \tau_l$: Binary thresholds for short and long intervals 
    \State $\mathcal{K}_\text{open}, \mathcal{K}_\text{close}$: Morphological kernels 
    \EndIndent

    \LineComment{Step 1: Global Motion Compensation}
    \State $H_{t, t-1} \coloneqq \text{RANSAC\_SIFT}(I_t, I_{t-1})$ \Comment{Estimate short-term $H$}
    \State $H_{t, t-5} \coloneqq \text{RANSAC\_SIFT}(I_t, I_{t-5})$ \Comment{Estimate long-term $H$}
    
    \State $\tilde{I}_{t-1} \coloneqq \text{WarpPerspective}(I_{t-1}, H_{t, t-1})$
    \State $\tilde{I}_{t-5} \coloneqq \text{WarpPerspective}(I_{t-5}, H_{t, t-5})$
    \State $R_{t-1} \coloneqq \text{WarpPerspective}(\mathbf{1}^{H \times W}, H_{t, t-1})$
    \State $R_{t-5} \coloneqq \text{WarpPerspective}(\mathbf{1}^{H \times W}, H_{t, t-5})$ 
    \State $\Delta_{t,1} \coloneqq |I_t - \tilde{I}_{t-1}| \odot R_{t-1}$ \Comment{Compute absolute differences}
    \State $\Delta_{t,5} \coloneqq |I_t - \tilde{I}_{t-5}| \odot R_{t-5}$

    \LineComment{Step 2: Asymmetric Morphological Filtering}
    \State $\Delta_s^{\text{blur}} \coloneqq \text{GaussianBlur}(\Delta_{t,1}, 5\times5)$ \Comment{Preserve fine structures}
    \State $D_s \coloneqq \text{Threshold}(\Delta_s^{\text{blur}}, \tau_s)$ 
    
    \State $D_l^{\text{raw}} \coloneqq \text{Threshold}(\Delta_{t,5}, \tau_l)$ 
    \State $D_l \coloneqq \text{MorphologyEx}(D_l^{\text{raw}}, \text{OPEN}, \mathcal{K}_\text{open})$ \Comment{Suppress isolated noise}

    \LineComment{Step 3: Spatiotemporal Fusion}
    \State $M_{\text{intersect}} \coloneqq D_s \text{ AND } D_l$ \Comment{Bitwise intersection}
    \State $M \coloneqq \text{MorphologyEx}(M_{\text{intersect}}, \text{CLOSE}, \mathcal{K}_\text{close})$

    \State \textbf{return} $M$

  \end{algorithmic}
\end{algorithm}

\textbf{Dual-Interval Difference Generation.} A single temporal interval is insufficient to balance noise suppression and accurate localization. Therefore, we extract motion masks at two complementary scales. The short-term mask $D_s$ (with $k=1$) utilizes a low threshold $\tau_s$ to capture the accurate instantaneous boundaries of targets, but it inevitably includes residual alignment noise.
To mitigate noise while preserving structures, the short-term difference is Gaussian smoothed by $\mathcal{G}_{5\times5}$ and binarized at $\tau_s$:
\begin{align}
    D_s = \mathcal{T}_{\tau_s}(\mathcal{G}_{5 \times 5}(\Delta_{t,1})). 
\end{align}
In contrast, the long-term mask $D_l$ (with $k=5$) emphasizes the accumulated displacement of targets. A stricter threshold $\tau_l$ ($\tau_l > \tau_s$) is applied, followed by a morphological opening operation $\mathcal{E}_\text{open}$ to eliminate isolated noise, calculated as
\begin{align}
    D_l = \mathcal{E}_\text{open}(\mathcal{T}_{\tau_l}(\Delta_{t,5})). 
\end{align}

\textbf{Spatiotemporal Fusion.} The final motion mask $M \in \{0,1\}^{H \times W}$ is obtained by computing the consensus (bitwise AND) of the two masks and refining the result via a morphological closing operation $\mathcal{E}_{close}$, given by
\begin{equation}
M = \mathcal{E}_{close}(D_s \wedge D_l).
\end{equation}

\begin{figure*}[t]
    \centering

    \includegraphics[trim=150pt 0 300pt 0, clip, width=0.24\textwidth]{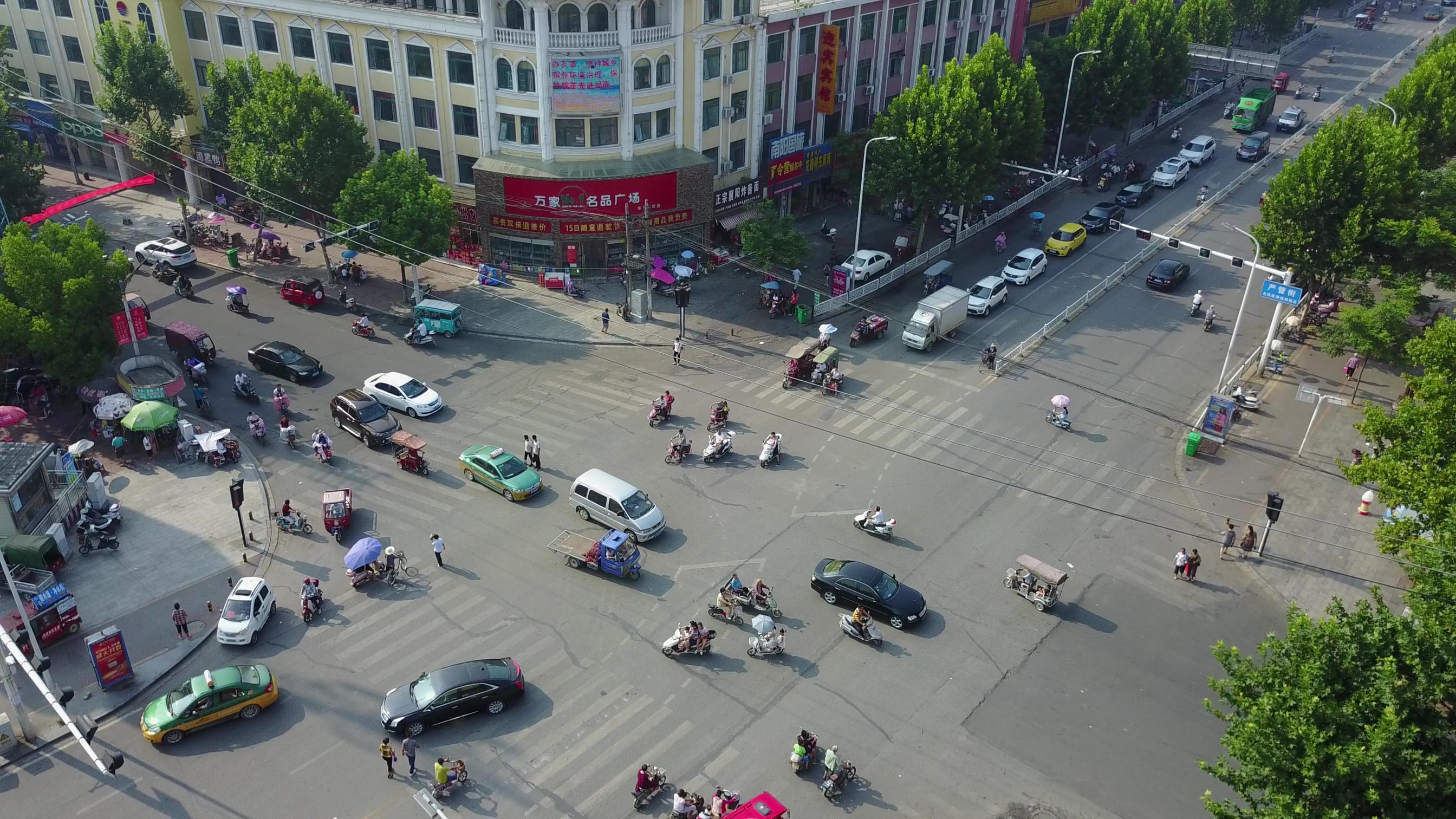}\hfill
    \includegraphics[trim=150pt 0 300pt 0, clip, width=0.24\textwidth]{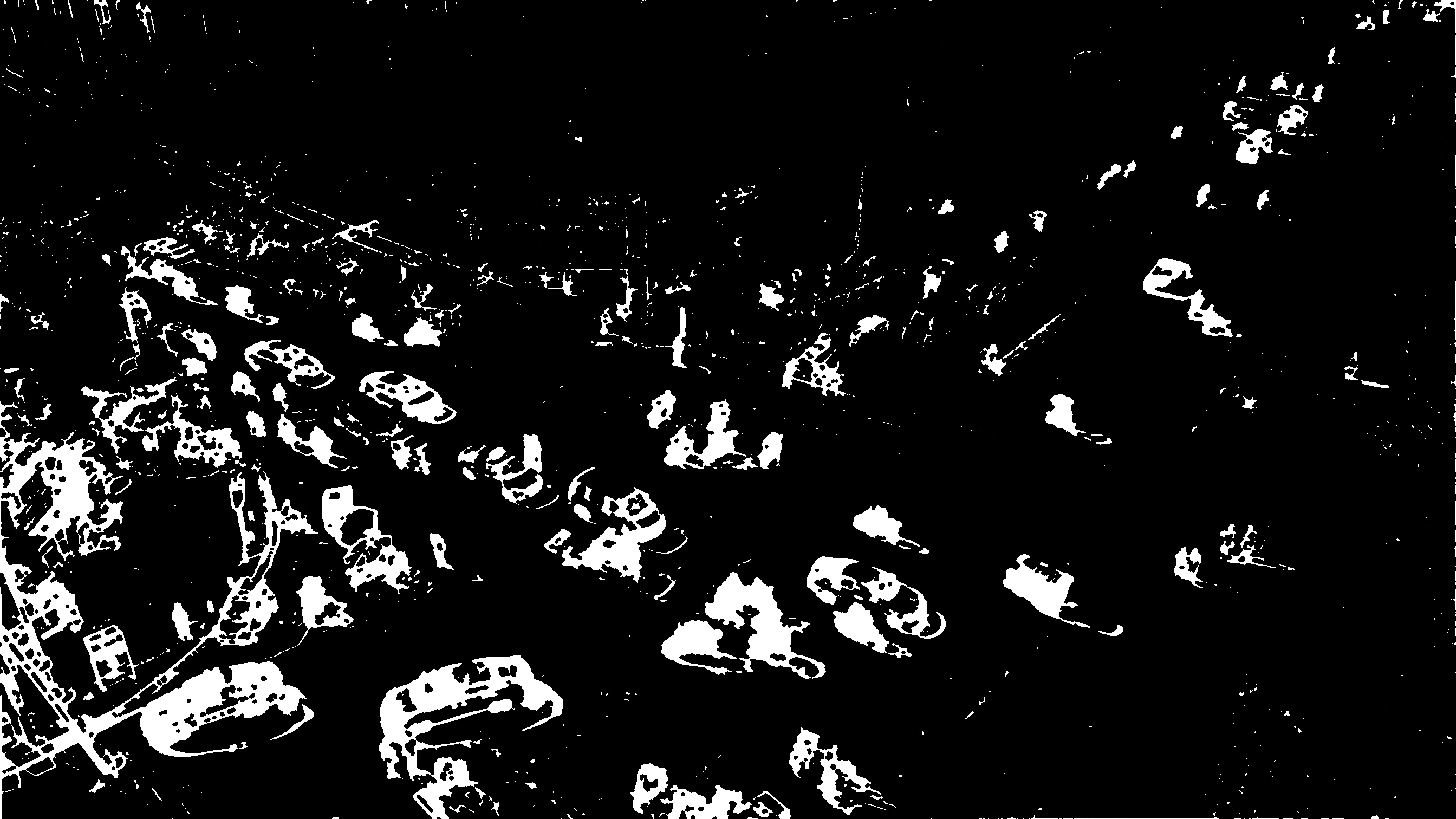}\hfill
    \includegraphics[trim=150pt 0 300pt 0, clip, width=0.24\textwidth]{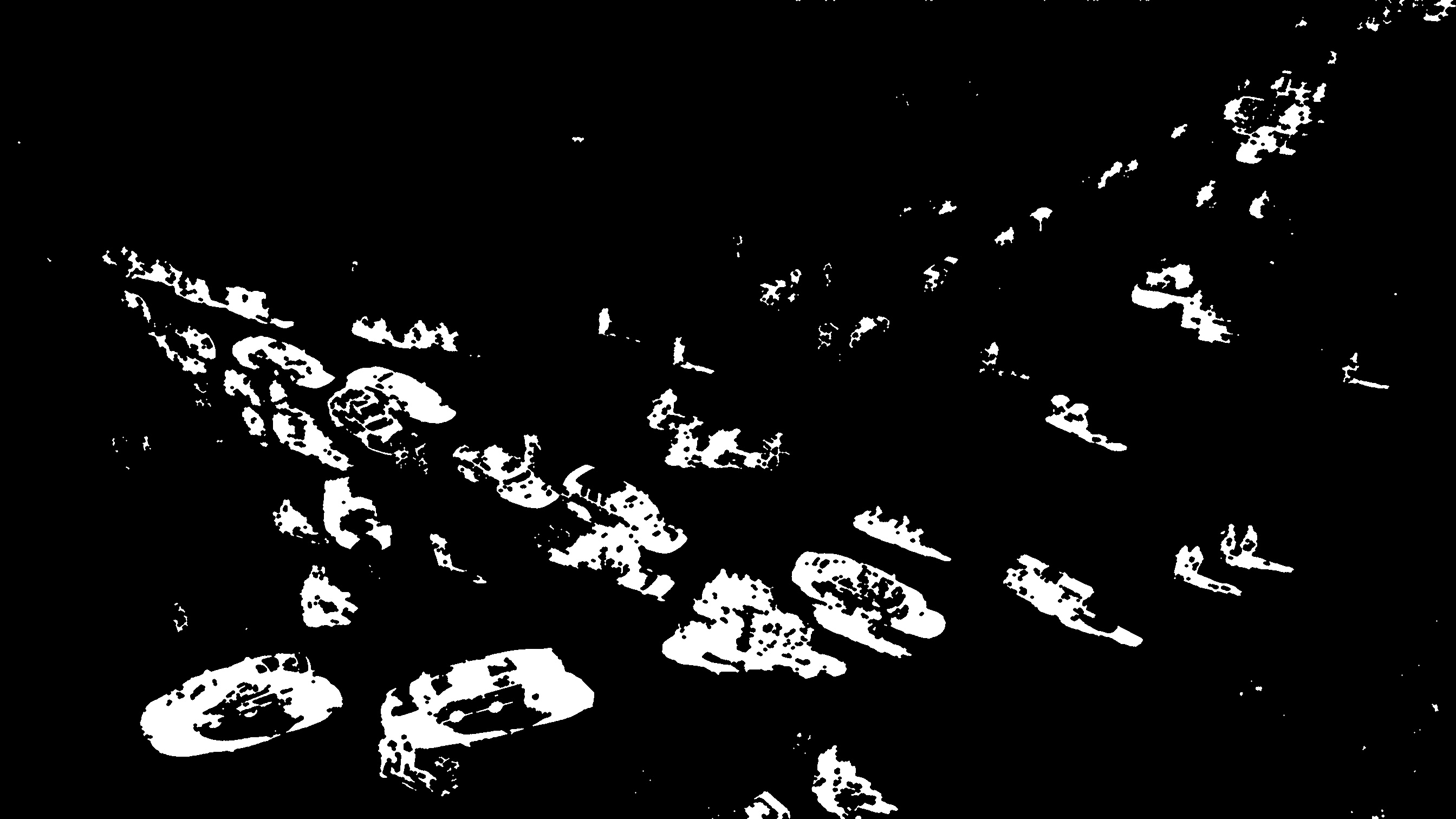}\hfill
    \includegraphics[trim=150pt 0 300pt 0, clip, width=0.24\textwidth]{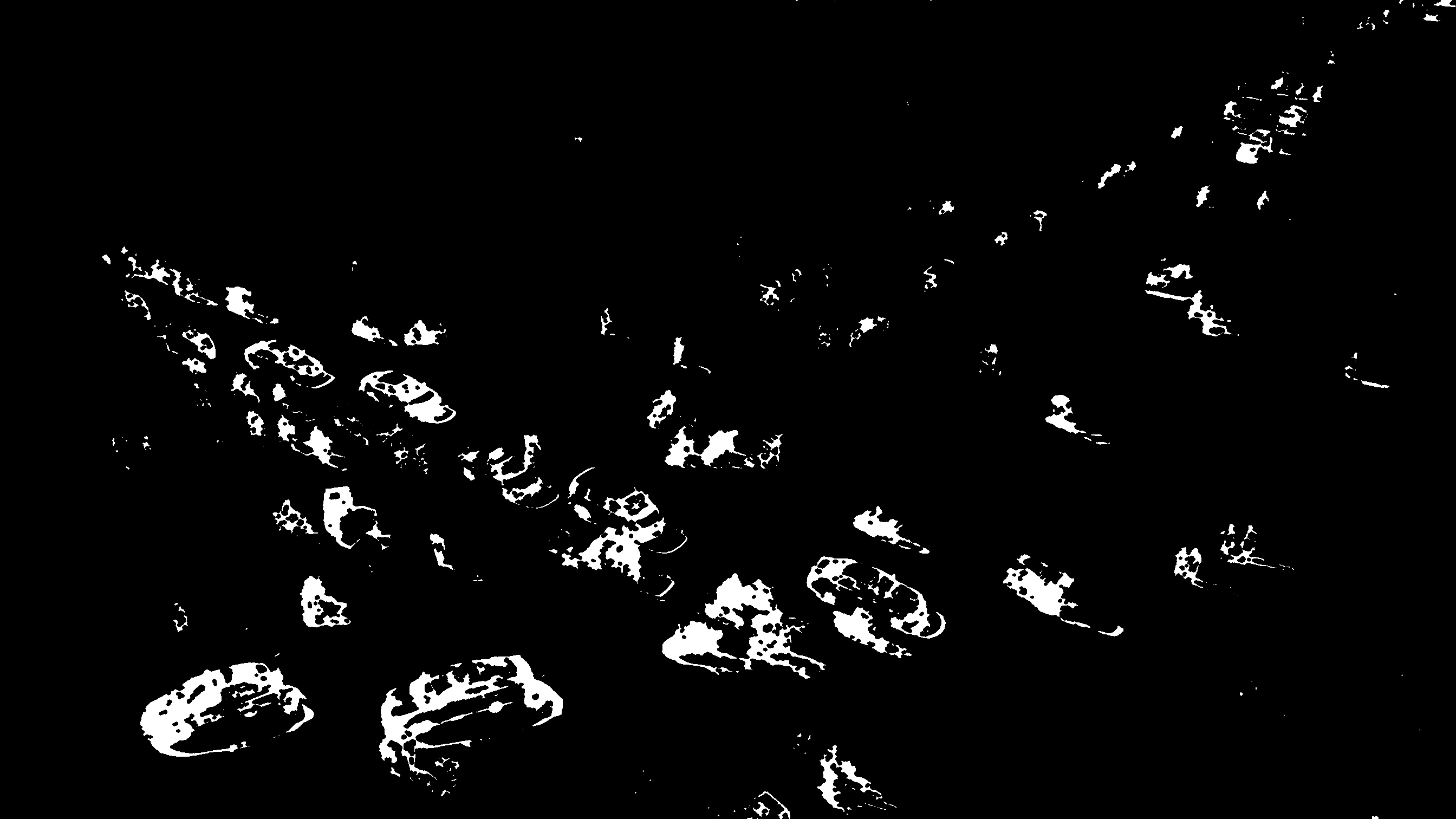}\\
    \makebox[0.24\textwidth]{\small (a) Original RGB Frame}\hfill
    \makebox[0.24\textwidth]{\small (b) Short-term Diff}\hfill
    \makebox[0.24\textwidth]{\small (c) Long-term Diff}\hfill
    \makebox[0.24\textwidth]{\small (d) Refined Motion Map}
    
    \vspace{0.05cm}
    
    \caption{Visualization of the proposed dual-interval motion extraction process. The short-term difference captures accurate instantaneous positions but is sensitive to ego-motion noise, whereas the long-term difference enhances motion saliency but introduces severe ghosting. The proposed fusion strategy acts as a mutual gate, utilizing long-term saliency to mask out short-term noise and short-term localization to trim long-term ghosting, producing a refined motion mask for subsequent detection.} 
    \label{fig:method_visualization}
\end{figure*}

Algorithm~\ref{alg:dual_interval} summarizes the offline motion-mask generation pipeline, with intermediate results visualized in Fig.~\ref{fig:method_visualization}. Short-term differencing provides responses close to the target's current position but remains sensitive to residual registration noise, whereas long-term differencing increases motion saliency but can introduce displaced responses from earlier target positions. Their intersection uses long-term support to filter short-term noise and short-term localization to restrict long-term ghosting, producing a refined motion mask. To reduce the cost of two independent feature-matching operations per frame, Sec.~\ref{network} introduces a lightweight implementation for real-time edge deployment.

\subsection{Dual-Interval Motion Injection (DIMI)}\label{secDIMI}

Although the extracted motion mask $M$ provides reliable localization of moving regions, it is inherently sparse and binary. The sparse nature of motion masks motivates direct guidance of multi-scale features rather than relying solely on propagation from the backbone input \cite{8765412, deepsurvey}. To address this issue, we propose a lightweight Dual-Interval Motion Injection (DIMI) module that utilizes the motion prior as soft spatial attention to amplify target features.

Given an intermediate feature map $F \in \mathbb{R}^{C \times H' \times W'}$, the motion mask $M$ is first resized to $H' \times W'$ via bilinear interpolation, yielding $M_{\downarrow}$. The precomputed mask $M_{\downarrow}$ guides soft spatial attention in the FPN through the Dual-Interval Motion Injection (DIMI) module. The attention map $A \in \mathbb{R}^{1 \times H' \times W'}$ is predicted strictly from the external mask $M_{\downarrow}$ rather than from the feature map $F$ and is given by
\begin{equation}
A = \sigma(\mathrm{BN}(\mathrm{Conv}_{3\times3}(M_{\downarrow}))),
\end{equation}
where $\sigma$ denotes the Sigmoid activation. The refined feature is obtained by residual modulation, applied as in the residual attention of Wang et al. {\cite{Wang2017}}:
\begin{equation}
F_{\mathrm{out}} = F \odot (1 + A),
\end{equation}
where \mbox{$(1+A)$} is broadcast over the \mbox{$C$} channels of \mbox{$F$}. This formulation scales the input features by motion-conditioned factors between one and two. Features with larger attention weights receive stronger amplification, while those with attention weights close to zero remain approximately unchanged.

\subsection{Network Architecture and Training Strategy}\label{network}

\textbf{Integration With YOLOv8.} The proposed framework is built upon the YOLOv8n architecture. DIMI is embedded at the P3, P4, and P5 levels of the YOLOv8n FPN, with strides of $8$, $16$, and $32$, respectively, to guide multi-scale features before the detection head.

The motion mask is injected exclusively into the FPN via the DIMI module (as shown in Fig. 2), while the YOLO backbone processes only the standard 3-channel RGB frame. 
However, to ensure that geometric data augmentations (e.g., Mosaic, Affine transformations) maintain strict spatial alignment between the RGB frame and the mask during training, the dataloader temporarily concatenates the RGB frame and the motion mask into a 4-channel tensor. Once the synchronous augmentations are applied, the tensor is decoupled into its respective 3-channel and 1-channel components prior to the network's forward pass.

\textbf{Padding During Geometric Augmentation}
When RGB and motion-mask channels are temporarily concatenated for synchronized geometric augmentation, applying the standard YOLO RGB padding value of 114 to all four channels introduces nonzero values into the mask borders. After normalization, these values become approximately 0.45, creating spurious motion cues in the padded regions.
To address this issue, RGB channels are padded with the standard value of $114$, while the motion channel is padded with $0$. This preserves the meaning of the motion mask without altering the RGB channels or introducing new visual features.

\begin{figure}
    \centering
    
    \includegraphics[width=0.8\linewidth]{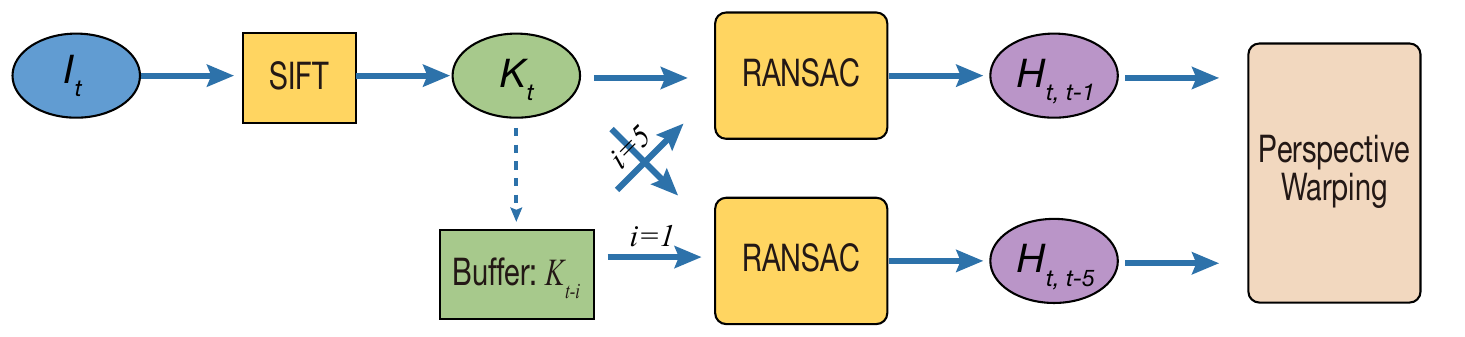}
    \makebox[0.24\textwidth]{\small (a) Offline Training Phase: Independent Matching}
    \includegraphics[width=0.8\linewidth]{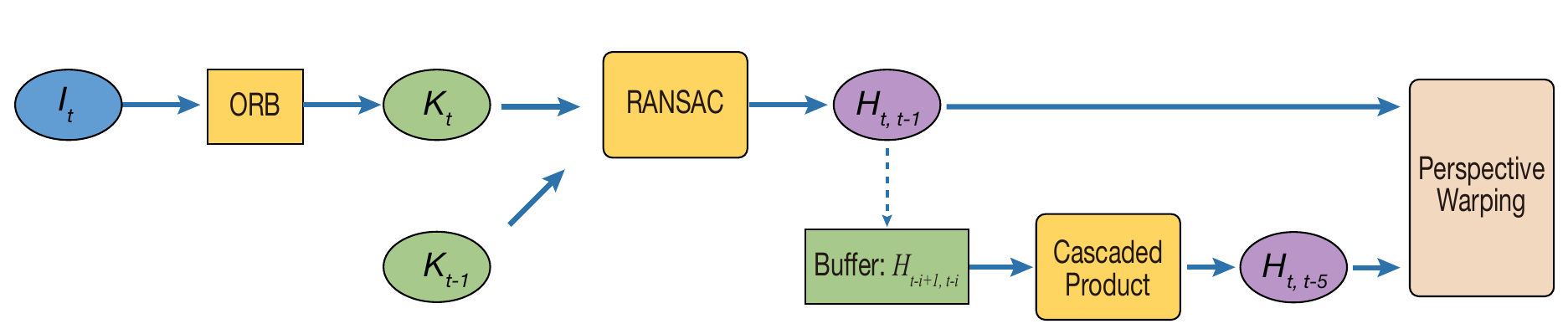}
    \makebox[0.24\textwidth]{\small (b) Online Inference Phase: Homography Cascading}
    \caption{Illustration of the proposed asymmetric training-inference paradigm for homography acquisition. During offline training, high-precision SIFT matching is independently performed for each interval to ensure accurate spatial alignment. During online inference, lightweight ORB matching and homography cascading are adopted to improve real-time deployment efficiency.} 
    \label{fig:asymmetric_paradigm}
\end{figure}

\textbf{Asymmetric Training-Inference Paradigm.} Figure~\ref{fig:asymmetric_paradigm} illustrates the SIFT-based offline and ORB-based online motion-extraction pipelines.
During the offline training phase, guaranteeing spatial alignment between the motion mask and RGB features is  paramount for training performance. Thus, we implement the motion extraction using high-precision SIFT descriptors with independent interval matching (as formulated in Sec. \ref{subsec3}) to generate spatial masks. However, the substantial computational overhead of independent feature matching across multiple intervals compromises the real-time capabilities of resource-constrained UAV platforms.

Therefore, we deploy a lightweight alternative using ORB \cite{2011ORB} features coupled with a homography cascading strategy. Based on the assumption of background rigidity between adjacent frames, the long-term transformation matrix $H_{t, t-k}$ (where $k=5$ in our implementation) is directly derived from a sliding window of historical short-term matrices via cascaded multiplication:
\begin{align}
    H_{t, t-k} = \prod_{i=1}^{k} H_{t-i+1, t-i}.
\end{align}
While per-frame ORB extraction and short-term matching remain the dominant computational bottlenecks, this strategy replaces an additional $\mathcal{O}(N\log N)$ long-interval match ($N$: keypoints per frame) by the $\mathcal{O}(k)$ product in Eq.~9 ($k=5$). 
The impact of this inference substitution on detection accuracy is evaluated in Sec.~IV-D.

\section{Experiments}

\subsection{Experimental Setup}

We evaluate the proposed method on the VisDrone-VID dataset \cite{VisDrone}, which contains dense small objects and camera ego-motion typical of UAV operations. The official VisDrone-VID split contains 56 training sequences (24,201 frames) and 7 validation sequences (2,846 frames). Training uses every fifth frame (4,841 images). Validation uses all 2,846 frames. Detection is scored per frame on ten classes: pedestrian, people, bicycle, car, van, truck, tricycle, awning-tricycle, bus, and motor. Results are not averaged over a sequence, and no tracking metric is reported. Following standard COCO evaluation protocols, we report $\text{AP}_{\mathbf{50}}$, the primary challenge metric $\text{AP}$ (averaged over IoU thresholds from $0.50$ to $0.95$), and small-object average precision $\text{AP}_{\mathbf{S}}$ (area $<32^2$ pixels).

The proposed framework is implemented based on YOLOv8n and trained from scratch for 200 epochs with a batch size of 8 and an input resolution of $640\times640$. Following the asymmetric paradigm in Sec.~\ref{network}, offline training masks are generated using SIFT-based independent matching. Training and accuracy evaluation are performed in PyTorch on an NVIDIA RTX 4090, while runtime is measured on the NVIDIA Jetson Orin Nano using TensorRT FP16 and the ORB motion front-end. Parameter counts and GFLOPs refer to the neural network, with GFLOPs computed at the same input resolution.

\begin{figure}[htbp]
    \centering
    \includegraphics[trim={3cm 20cm 3cm 1cm}, clip, width=0.48\textwidth]{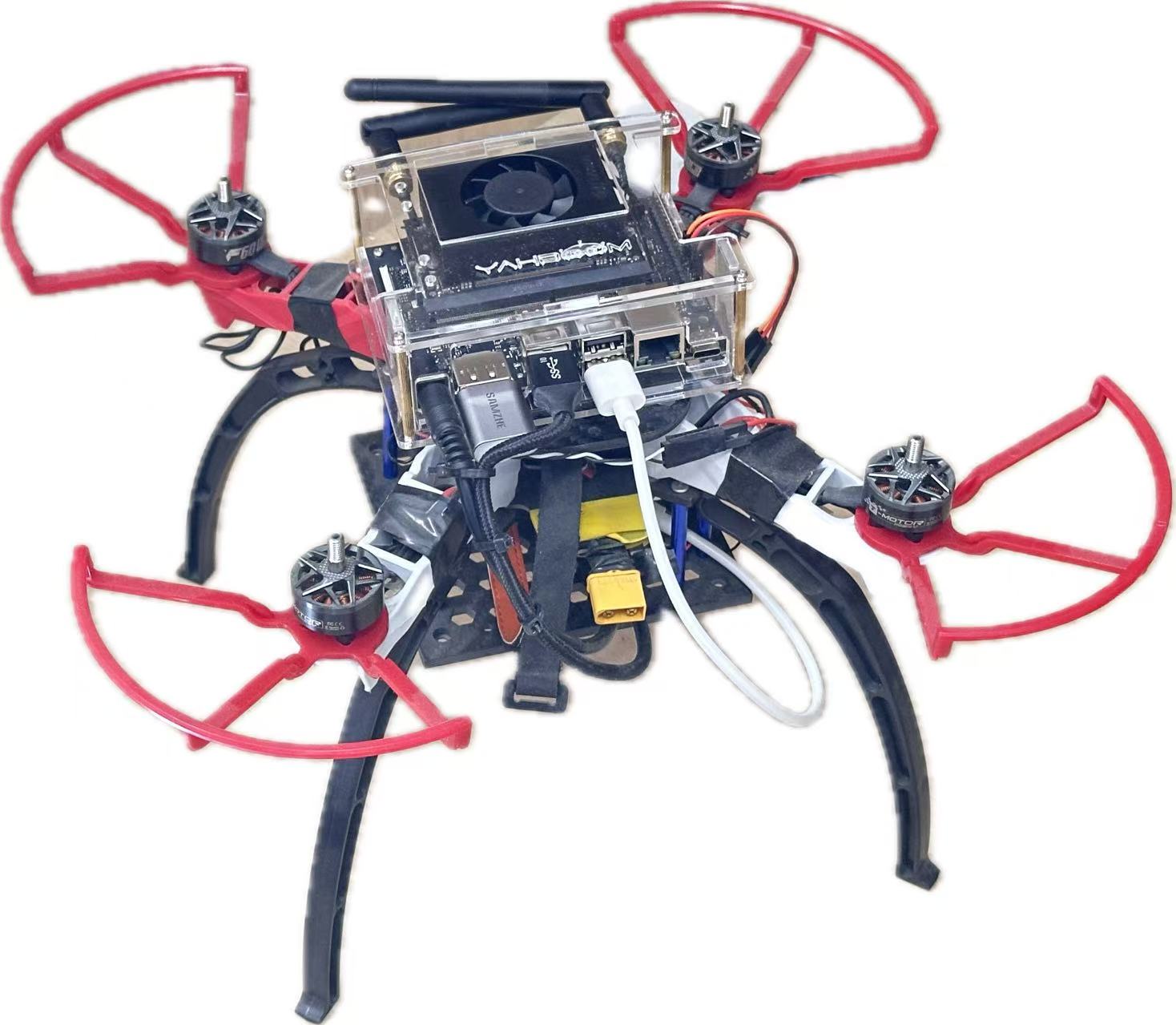}
    \caption{Deployment platform used for real-time edge inference experiments based on the NVIDIA Jetson Orin Nano.}
    \label{fig:nano}
\end{figure}

\subsection{Comparison with State-of-the-Art Methods}

\begin{table*}[!t]
\centering
\caption{Comparison on the VisDrone-VID validation set at $640\times640$. Accuracy is evaluated in desktop PyTorch, while detector-only and end-to-end speeds are measured on the NVIDIA Jetson Orin Nano using TensorRT FP16.}
\label{tab:sota}
\footnotesize
\setlength{\tabcolsep}{4pt}
    \centering
    \footnotesize
    \begin{tabular*}{0.9\textwidth}{@{\extracolsep{\fill}}lcccccccc@{}}
        \toprule
        \textbf{Method} & \textbf{Init} & \textbf{Params} & \textbf{GFLOPs}  & \textbf{AP$_{\mathbf{50}}$} & \textbf{AP} &\textbf{AP$_{\mathbf{S}}$} & \textbf{$\text{FPS}_{\text{det}}$} & \textbf{$\text{FPS}_{\text{e2e}}$}\\
        \midrule
        \textit{Static Detectors} & & & & & & \\
        RT-DETR-L     & COCO & 32.0\,M & 103.5 & 30.3 & 15.5 & 5.89 & 36.2 & 18.0 \\
        YOLOv8s       & COCO & 11.1\,M & 28.5  & 24.9 & 12.8 & 3.99 & 45.7 & 27.7 \\
        YOLO11n       & COCO & 2.58\,M & 6.3 & 22.9 & 11.0 & 3.28 & 73.9 & 38.3 \\
        YOLOv8n (Baseline)\,$\dagger$ & scratch & 3.01\,M  & 8.1 & 22.8 & 9.7 & 2.94 & 82.8 & 40.8 \\
        \midrule
        \textit{Adapted Motion-Fusion} & & & & & & \\
        Multi-frame RGB Input $\dagger$  & scratch & 3.01\,M  & 8.2  & 24.2 & 10.8 & 3.77 & 73.4 & 33.7 \\
        Early Mask Fusion $\dagger$ & scratch & 3.01\,M  & 8.1  & 23.6 & 10.8 & 3.32 & 75.7 & 21.9 \\
        \midrule
        \textit{Video-Aware Baselines} & & & & & & \\
        FGFA        & ImageNet & 103.2\,M  & 460.7  & 17.7 & 7.1 & 0.90 & $-$ & $-$ \\
        YOLOFT-s  & COCO & 13.9\,M  & 38.4  & 23.3 & 11.4 & 4.97 & 15.9 & 13.1 \\
        \midrule
        \textbf{Ours}\,$\dagger$  & scratch & \textbf{3.01\,M}  & \textbf{8.1} & \textbf{25.8} & \textbf{11.4} & \textbf{3.36} & \textbf{75.8} & \textbf{36.8}\\
        \bottomrule
      \end{tabular*}
    \par\smallskip
    \begin{minipage}{0.95\textwidth}
        \footnotesize
        Init denotes the initialization source. $\dagger$ marks models trained from scratch under the same training schedule, with results averaged over three independent runs using different random seeds. Other methods use their respective official configurations and report single-run results. $\mathrm{FPS}_{\mathrm{det}}$ measures network forward-pass speed. $\mathrm{FPS}_{\mathrm{e2e}}$ includes preprocessing, network inference, NMS, and motion extraction where applicable. Params and GFLOPs refer to the neural network. For our framework, end-to-end speed uses the parallel schedule.
    \end{minipage}
\end{table*}

To evaluate the effectiveness and efficiency of the proposed framework, we compare it against existing state-of-the-art object detectors on the VisDrone-VID validation set. The evaluated models include the baseline YOLOv8n, the scaled-up YOLOv8s \cite{yolov8_ultralytics}, the YOLO11n \cite{yolo11_ultralytics}, and the Transformer-based RT-DETR-L \cite{RT-DETR-L}.

We include the Multi-frame RGB Input baseline that concatenates $I_t$ and $I_{t-5}$ along the channel axis and feeds them to YOLOv8n without GMC or an explicit motion mask, and an Early Mask Fusion baseline that concatenates the same GMC mask with RGB at the backbone input, inspired by the early-fusion approach of MG-VTOD {\cite{yang2023video}}. Together with FPN concatenation, these variants compare alternative mask entry points and fusion operators.

Furthermore, we include dedicated video-aware baselines such as FGFA {\cite{fgfa}} and YOLOFT {\cite{yoloft}}. FGFA and YOLOFT-s are trained on the VisDrone-VID training split at the shared 640 input resolution for protocol parity. FGFA uses an ImageNet-initialized ResNet-50 and a pretrained FlowNet, whereas YOLOFT-s initializes compatible layers from COCO-pretrained YOLOv8s weights. FGFA's GFLOPs are counted per frame with 30 reference frames. Its mmcv implementation does not run on the Jetson, so no FPS is reported.

As shown in Table~\ref{tab:sota}, the proposed framework achieves an AP$_{\mathbf{50}}$ of 25.8 and an AP of 11.4, improving upon the baseline YOLOv8n by 3.0 and 1.7 points, respectively. The proposed detector contains 3,012,831 parameters, compared with 3,012,798 for the YOLOv8n baseline. DIMI therefore adds 33 learnable parameters and less than 0.002 GFLOPs at an input resolution of $640\times640$.
For the proposed framework the detector runs at 75.8 FPS, while the full pipeline, including GMC and dual-interval extraction, runs at 36.8 FPS (27.17\,ms) on the Jetson Orin Nano. The runtime breakdown and parallel schedule are detailed in Section~\mbox{\ref{sec_profiling}}.

Compared to larger architectures, the proposed method provides a balance between accuracy and computational cost. While RT-DETR-L achieves an AP$_{\mathbf{50}}$ score of 30.3, this higher accuracy requires substantially more computation. The RT-DETR-L model requires 32.0 M parameters and 103.5 GFLOPs, with detector-only and end-to-end speeds of 36.2 and 18.0 FPS, respectively.
At the shared input resolution of $640\times640$, FGFA achieves 17.7 $\text{AP}_{\mathbf{50}}$. YOLOFT-s achieves 23.3 $\text{AP}_{\mathbf{50}}$ at 13.1 $\text{FPS}_{\text{e2e}}$ with 13.9\,M parameters and 38.4 GFLOPs.

\begin{figure*}[!t]
    \centering

    \includegraphics[width=0.33\linewidth]{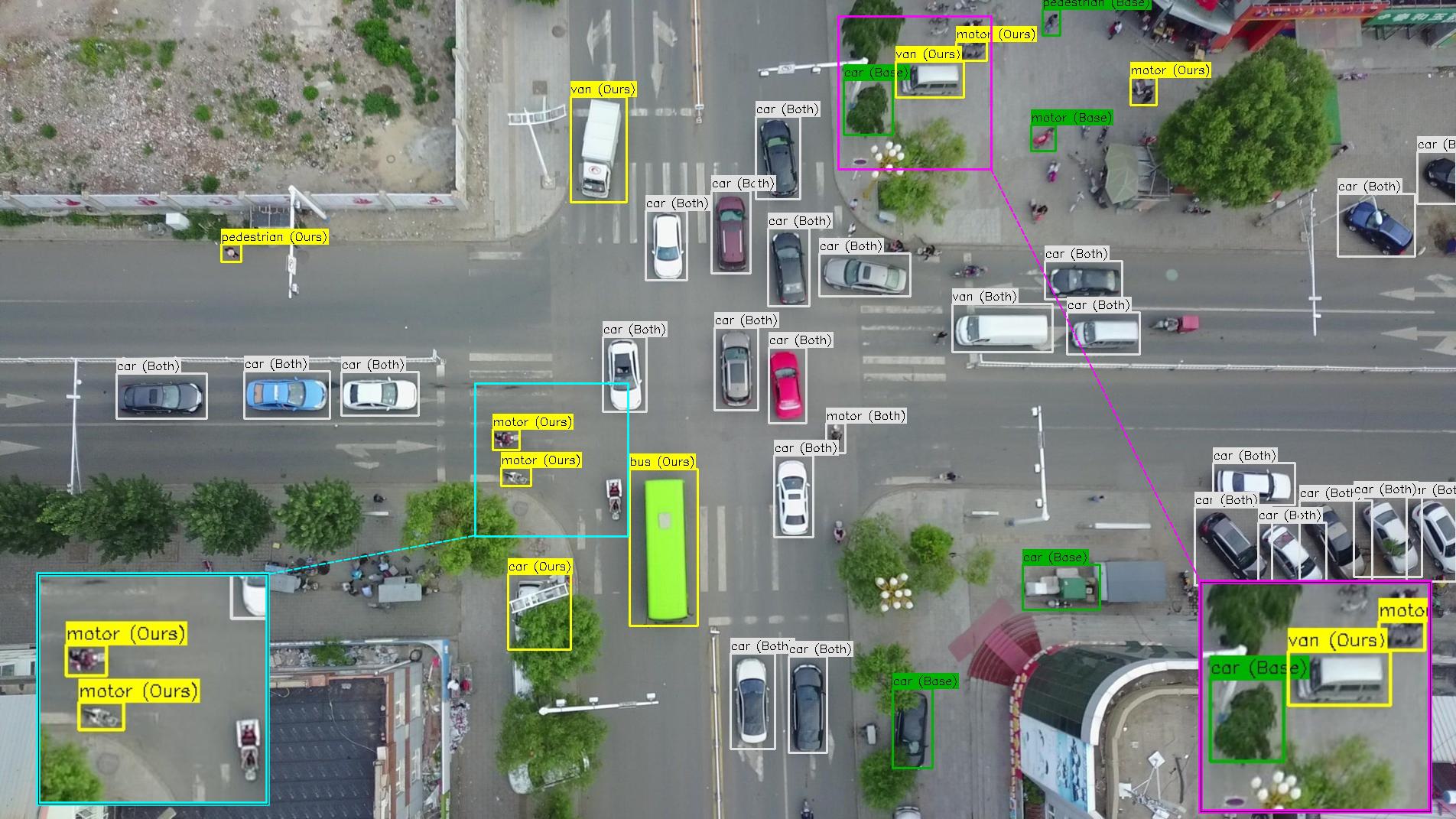}\hfill
    \includegraphics[width=0.33\linewidth]{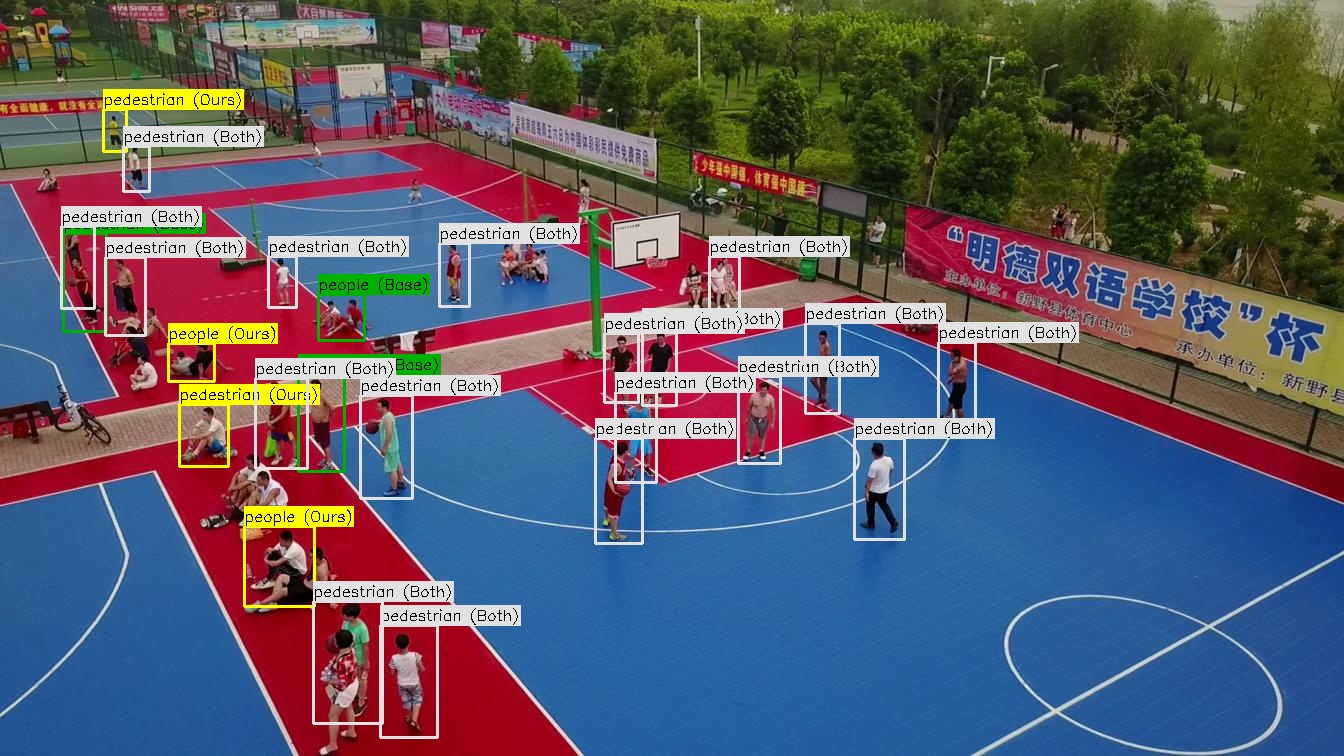}\hfill
    \includegraphics[width=0.33\linewidth]{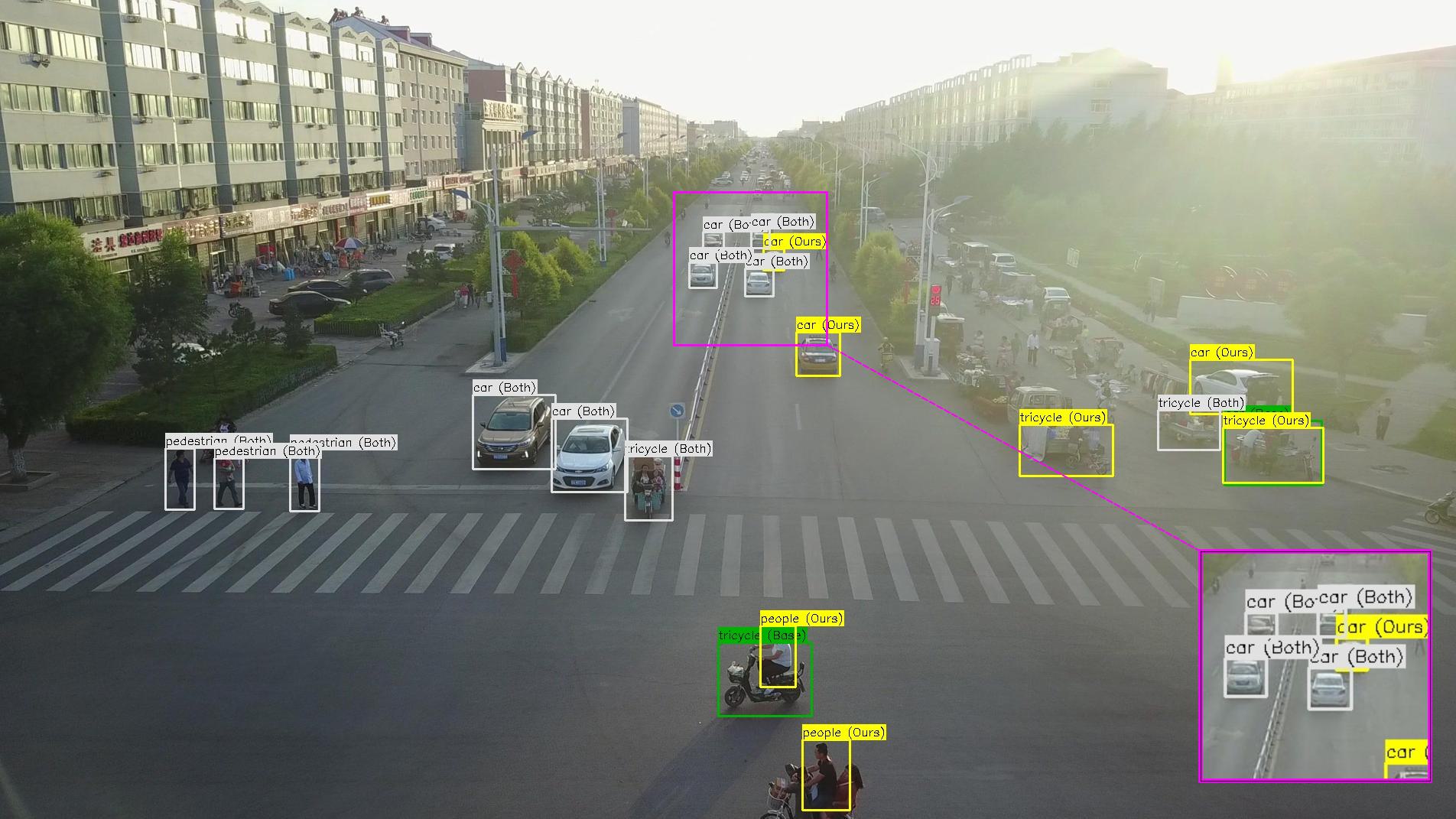}\hfill
    \vspace{0.05cm} 

    \includegraphics[width=0.33\linewidth]{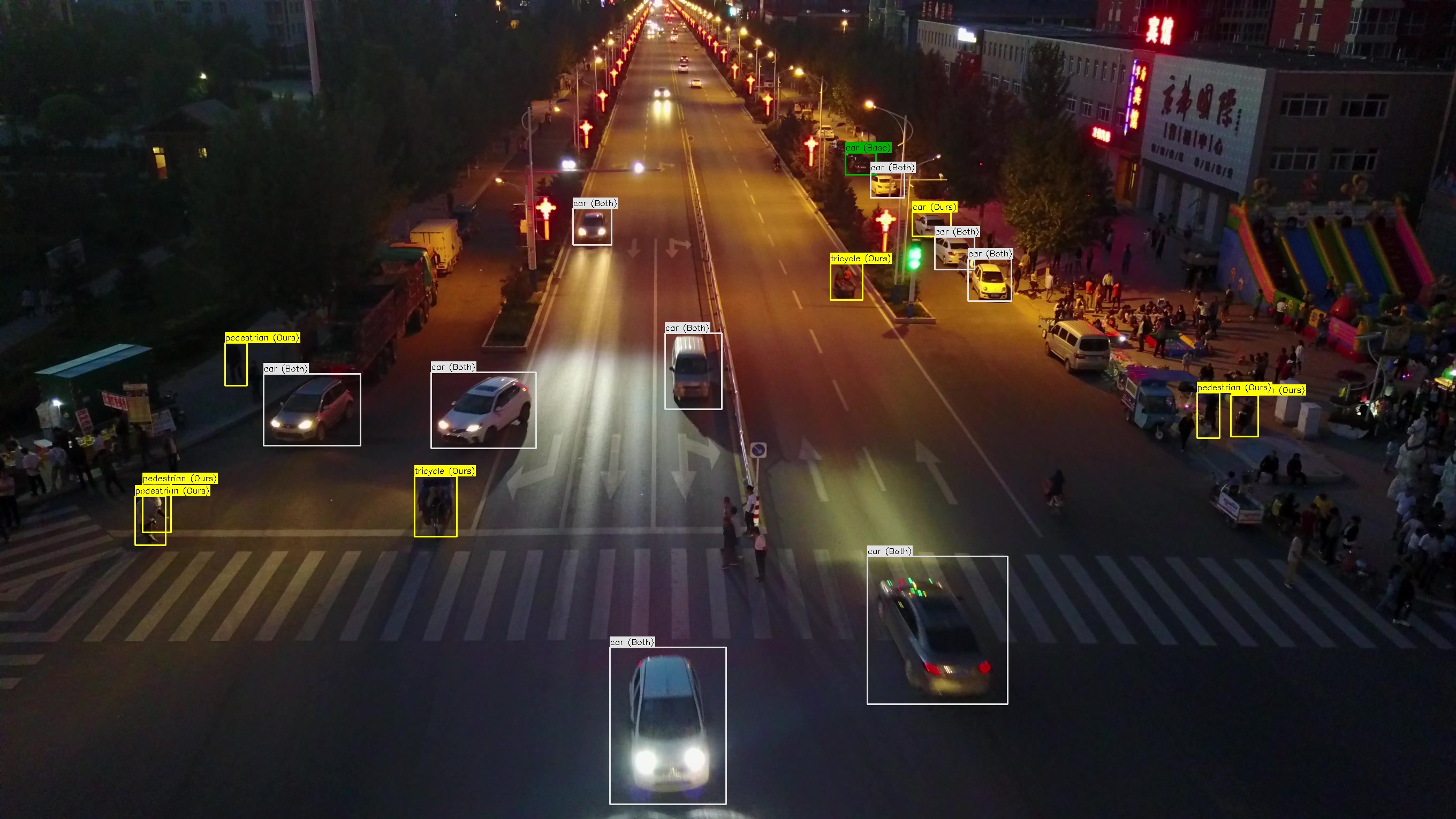}\hfill
    \includegraphics[width=0.33\linewidth]{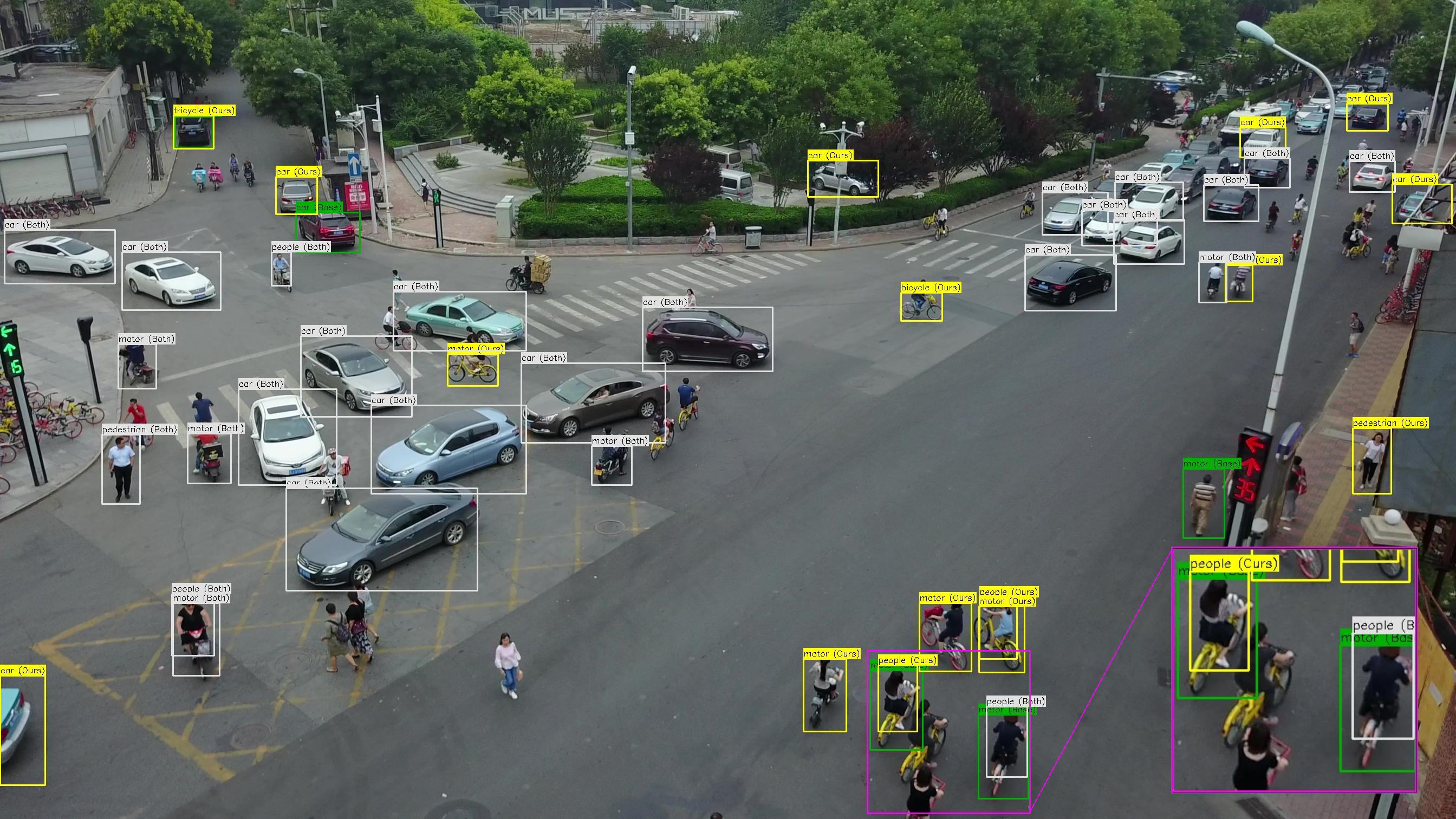}\hfill
    \includegraphics[width=0.33\linewidth]{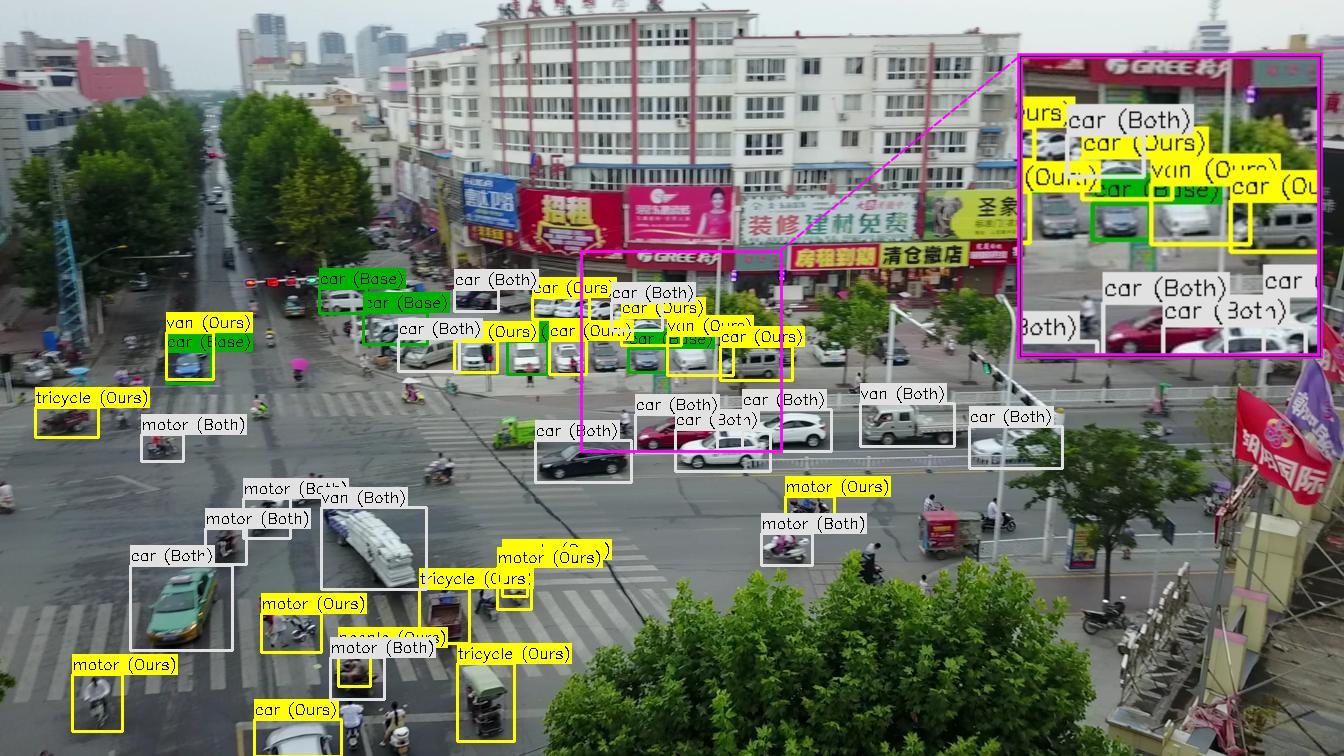}\hfill
    \vspace{0.05cm} 
    
    \caption{Qualitative comparison between the spatial-only baseline and the proposed framework on the VisDrone-VID validation set. Yellow boxes denote detections unique to the proposed method, green boxes indicate detections unique to the baseline, and white boxes represent detections shared by both models. The proposed motion-guided framework improves robustness against occlusion, cluttered backgrounds, and small-object ambiguity.}
    \label{fig:qualitative_frames}
\end{figure*}

\subsection{Ablation Study}

\begin{table}[htbp]
    \centering
    \caption{Ablation study of the proposed components on the VisDrone-VID validation set. All variants are trained from scratch in our codebase under the identical protocol, and results are reported as the mean $\pm$ population standard deviation over three independent runs with different random seeds. N/A denotes absence of a mask channel.}
    \label{tab:ablation}
    \resizebox{0.48\textwidth}{!}{
    \begin{tabular}{c c c c | c c}
        \toprule
        \multicolumn{2}{c}{\textbf{Dual-Interval Extraction}} & \multirow{2}{*}{\textbf{Fusion Module}} & \multirow{2}{*}{\textbf{Mixed Padding}} & \multirow{2}{*}{\textbf{AP$_{\mathbf{50}}$}} & \multirow{2}{*}{\textbf{AP}} \\
        \makecell{Short-term \\ ($\Delta t=1$)} & \makecell{Long-term \\ ($\Delta t=5$)} & & & & \\
        \midrule
        \multicolumn{6}{l}{\textit{Part 1: Component and Padding Analysis}} \\
        \ding{55} & \ding{55} & \ding{55} & N/A (Baseline) & 22.8 $\pm$ 1.2& 9.7 $\pm$ 0.7 \\
        \ding{51} & \ding{51} & DIMI & \ding{55} & 23.9 $\pm$ 0.5 & 10.6 $\pm$ 0.3 \\
        \ding{51} & \ding{51} & DIMI & \ding{51} (Ours) & \textbf{25.8 $\pm$ 1.5} & \textbf{11.4 $\pm$ 0.5} \\
        \midrule
        \multicolumn{6}{l}{\textit{Part 2: Motion Extraction and Fusion Variants}} \\
        \ding{51} & \ding{55} & DIMI & \ding{51} & 24.7 $\pm$ 0.2 & 11.1 $\pm$ 0.3 \\
        \ding{55} & \ding{51} & DIMI & \ding{51} & 23.3 $\pm$ 1.0 & 10.2 $\pm$ 0.5 \\
        \ding{51} & \ding{51} & Concat & \ding{51} & 24.1 $\pm$ 0.8 & 10.7 $\pm$ 0.6 \\
        \bottomrule
    \end{tabular}
    }
\end{table}

Table~\ref{tab:ablation} summarizes the ablation results. The RGB-only YOLOv8n baseline achieves $22.8\,\mathrm{AP}_{50}$. 
Adding short-interval or long-interval motion cues yields 24.7 and 23.3, respectively. Combining both intervals achieves the highest mean \mbox{$\mathrm{AP}_{50}$} among the evaluated variants at 25.8, supporting the complementary roles of short-term localization and long-term motion saliency. FPN concatenation reaches \mbox{$24.1\,\mathrm{AP}_{50}$}, compared with 25.8 for DIMI's soft spatial attention.

To further evaluate the effectiveness of the proposed motion-guided framework, we conduct a qualitative comparison against the baseline on the VisDrone-VID validation set, as illustrated in Fig.~\ref{fig:qualitative_frames}.

To quantify the detection discrepancy, each ground-truth box on the validation set is matched to a detection of the same class at IoU 0.5, using a confidence threshold of 0.4. Of the ground-truth boxes hit by the baseline, 80.31\% are also hit by our model and 19.69\% are not. Of the ground-truth boxes missed by the baseline, our model hits 6.24\%. At this operating point, the baseline hits 28,280 ground-truth boxes and our model hits 28,051. Although the hit count is slightly lower at this threshold, our method achieves higher AP$_{50}$ when evaluated across confidence thresholds.


\subsection{Mask Padding and Runtime Analysis}\label{sec_profiling}

\textbf{Effect of Mask Padding.} The second and third rows of Table~\ref{tab:ablation} show the effect of the mask-border fill value. Applying the RGB padding value of 114 to the motion-mask channel introduces spurious motion cues at image boundaries. With this fill value, the full model achieves an AP$_{\mathbf{50}}$ of 23.9. Setting the mask-border fill to 0 prevents these artificial responses and increases AP$_{\mathbf{50}}$ to 25.8.

\textbf{Runtime and Asymmetric Inference Analysis.} On the NVIDIA Jetson Orin Nano, CPU preprocessing (including ORB mask extraction) takes 16.17\,ms, the TensorRT network forward pass takes 13.20\,ms (75.8 FPS), and NMS post-processing takes 2.67\,ms. Serial execution therefore takes 32.04\,ms per frame, corresponding to 31.2 FPS.
Because the DIMI module injects the motion mask at the FPN levels (P3, P4, P5) rather than at the input stem, the GPU-based RGB backbone can run concurrently with CPU-based ORB extraction for the same frame, hiding the backbone latency behind the CPU preprocessing window. 
Overlapping these stages reduces the measured end-to-end latency of our complete framework to 27.17\,ms, yielding the full-pipeline speed of 36.8 FPS reported in Table~\mbox{\ref{tab:sota}}.

On VisDrone-VID validation frames, the cascaded ORB mask covers 78.8\% of the SIFT foreground, with a micro IoU of 0.1574. Training and inferring with ORB reaches 23.6 AP$_{\mathbf{50}}$. Training on SIFT and inferring with cascaded ORB reaches 25.8, against 26.2 when SIFT is also used at inference. Despite the low pixel-level overlap, the SIFT-trained detector tolerates the mask substitution, with AP$_{\mathbf{50}}$ decreasing by only $0.4$ when using cascaded ORB at inference. Compared with the SIFT-based implementation (135.27 ms), the proposed ORB-based homography cascading strategy significantly improves deployment efficiency while preserving detection accuracy. These results demonstrate the feasibility of the proposed framework for real-time UAV perception on resource-constrained edge platforms.

\section{Conclusions}

In this paper, we proposed a Dual-Interval Motion-Guided Object Detection framework for UAV videos with camera ego-motion and complex backgrounds. Homography-based GMC aligns the background, while dual-interval differencing combines short-term localization with long-term motion saliency to refine the motion prior. The extracted motion cues were seamlessly integrated into the detection pipeline via the proposed Dual-Interval Motion Injection (DIMI) module. An asymmetric implementation uses SIFT-generated masks for training and cascaded ORB masks for inference. On VisDrone-VID, the framework achieves $25.8\,\mathrm{AP}_{50}$ with a 3.01 M-parameter detector, while reaching 36.8 FPS end-to-end on the Jetson Orin Nano. These results demonstrate the feasibility of incorporating explicit motion cues into a lightweight detector for real-time UAV video perception.

Despite the improved accuracy, the proposed framework relies on a global homography assumption for background motion compensation. During low-altitude flights with strong parallax or severe depth discontinuities, the 2D homography alignment degrades, producing residual artifacts. 
Strong parallax can corrupt the motion prior and affect downstream detection. Future work will explore integrating depth-aware monocular estimation or local optical flow to handle complex 3D perspective distortions.

\textbf{Code Availability:} To ensure reproducibility, the source code, the train and validation frame lists, the precomputed dual-interval masks, and the trained weights will be released.

\bibliographystyle{IEEEtran}
\bibliography{IEEEcontrol,ref}

\end{document}